\documentclass{article}

\PassOptionsToPackage{numbers,sort&compress}{natbib}
\usepackage[final,sglblindworkshop]{neurips_2026}
\workshoptitle{Foundation Models for the Brain and Body}

\usepackage[utf8]{inputenc}
\usepackage[T1]{fontenc}
\usepackage{hyperref}
\usepackage{url}
\usepackage{booktabs}
\usepackage{amsfonts}
\usepackage{amsmath}
\usepackage{nicefrac}
\usepackage{microtype}
\usepackage{xcolor}
\usepackage{graphicx}

\title{PACE-ECG: Patient-Aware Contrastive ECG Pretraining; Leveraging Longitudinal Structure for Transferable Representations}

\author{%
  \textbf{Riccardo Cavarra \quad Lupo Lovatelli \quad Shaheim Ogbomo-Harmitt} \\
  \textbf{Jakub Grzelak \quad Adelaide De Vecchi \quad Andrew King \quad Oleg Aslanidi} \\
  King's College London, St Thomas' Hospital, London, UK \\
  {\footnotesize\texttt{\{riccardo.cavarra, lupo.lovatelli, shaheim.ogbomo-harmitt,}} \\
  {\footnotesize\texttt{jakub.grzelak, adelaide.de\_vecchi, andrew.king, oleg.aslanidi\}@kcl.ac.uk}}
}

\begin{document}

\maketitle

\begin{abstract}
Foundation models for physiological signals are commonly evaluated on in-domain interpretation tasks, but their ability to transfer to longitudinal clinical prognosis across healthcare systems remains uncertain. We developed PACE-ECG, an ECG pretraining framework that combines contrastive learning across temporally neighbouring recordings from the same patient with clinically guided multitask supervision. We evaluated frozen ECG representations for post-myocardial-infarction all-cause mortality and incident heart failure prediction across MIMIC-IV and two external HEEDB cohorts, comparing the proposed representation with DeepECG-SSL, a state-of-the-art model, and two control baselines. Across both downstream tasks, the proposed model outperformed DeepECG-SSL in-domain (AUROC +1.9\% for mortality and +4.0\% for heart failure) and on external populations in HEEDB-Boston (+5.6\% for mortality +10.3\% for heart failure) and HEEDB-Atlanta (+6.9\% for mortality +11.0\% for heart failure) while using an encoder 41 times smaller and a fraction of the pretraining data. A matched ablation showed that patient-temporal pairing improved external mortality and heart failure AUROC over generic self-supervision by 7.0\% and 9.4\% on HEEDB-Boston and 8.9\% and 9.3\% on HEEDB-Atlanta. These findings indicate that pretraining-objective design greatly influences downstream transfer alongside model and data scale, and that in-domain performance alone is insufficient to establish the clinical generalisability of ECG foundation representations.
\end{abstract}

\section{Introduction}

Deep learning is increasingly used in medicine for applications ranging from diagnosis to risk prediction \cite{ref1}, yet most medical artificial intelligence remains built around narrow, task-specific models. Such approaches depend on large quantities of labelled data and require retraining when transferred to new tasks or data distributions.  Foundation models address this issue though self-supervised learning (SSL). These models are pretrained on large, unlabelled datasets and produce reusable representations and can support multiple downstream applications with limited task-specific supervision. This is particularly attractive in medicine, where expert annotation is scarce and expensive and data distributions vary across patient populations, institutions, acquisition systems and clinical workflows \cite{ref2}. The electrocardiogram (ECG) is an inexpensive, widely available measurement of cardiac electrophysiology, and large clinical archives make it a natural setting for representation learning.

\subsection{ECG representation learning}

A growing literature has explored ECG self-supervision through instance discrimination, latent forecasting, reconstruction and predictive embedding objectives \cite{ref3,mehari2022}. Masked transformers \cite{ref4}, joint-embedding predictive architectures \cite{ref5} and lead-agnostic local--global learning \cite{ref6} have demonstrated the potential of these approaches to produce transferable ECG representations.  Representation quality, however, depends on the pretraining objective as well as data scale, because the objective shapes which features and invariances the encoder preserves for downstream use. In contrastive learning, representations of positive pairs are drawn together, while those of negative pairs are pushed apart. The construction of these pairs therefore determines which variations the encoder is encouraged to preserve or disregard.  \cite{ref8}. 

ECG contrastive methods exploit several sources of shared information, including relationships across leads  \cite{chen2025tssl}, temporal segments \cite{clocs2021}, and physiologically motivated views generated through vectorcardiographic transformations and temporal masking  \cite{gopal2021}.
Patient identity extends contrastive learning beyond different views of a single recording. PCLR, for example, treats distinct recordings from the same patient as positive pairs, encouraging representations to capture information shared across acquisitions  \cite{diamant2022}. Related approaches combine subject-specific contrastive losses with adversarial learning to promote subject-invariant representations \cite{cheng2020}, integrate intra-subject temporal modelling with inter-subject contrastive learning \cite{lan2022}, or extend patient-aware learning to ECG images \cite{sangha2024}. More recent work combines patient-aware sampling with physiologically inspired augmentations and lead-based contrastive views \cite{dade2026}. These approaches differ in the information they seek to preserve, the unit of positive-pair construction and the temporal relationships they exploit.

The evaluation setting is equally important. DeepECG-SSL combines Wav2Vec, contrastive multi-segment coding and random lead masking, achieving strong ECG interpretation performance across multiple external healthcare datasets and transfer to selected tasks \cite{ref7}. These advances motivate a broader question: whether representations that perform well on conventional ECG benchmarks retain clinically meaningful prognostic information when transferred to independent healthcare systems. Addressing this question requires task- and cohort-specific evaluation of the information preserved by different pretraining schemes. 

\subsection{Study focus and contributions}

We investigate this question in the context of myocardial infarction (MI), a leading cause of death and disability worldwide, where the ECG is central to acute assessment but is underused for longer-term risk stratification \cite{ref9}. Mortality and incident heart failure are important outcomes: reported estimates include one-year mortality of 20\% and heart-failure occurrence of 20--30\%, with heart failure associated with a three-fold increase in mortality risk \cite{ref10,ref11}. Building on Cavarra et al. \cite{ref12}, we propose PACE-ECG, a pretraining framework which combines positive pairs of distinct same-patient ECGs acquired within 60 days with auxiliary waveform- and report-derived supervision. The pairing rule defines a local acquisition-time neighbourhood, while the auxiliary tasks encourage the encoder to retain morphological and rhythm information. We evaluate the resulting representations across MIMIC-IV and two external HEEDB cohorts, training prediction heads only on MIMIC-IV and applying them externally without adaptation.

Our contributions are twofold:
\begin{enumerate}
    \setlength{\itemsep}{0pt}
    \setlength{\parskip}{0pt}
    \setlength{\parsep}{0pt}
    \item We compare bounded same-patient pairing with generic same-ECG contrastive learning within a shared multitask pretraining framework, assessing the pairing scheme conditional on auxiliary supervision.
    \item We benchmark PACE-ECG, a generic SSL model and DeepECG-SSL representations for post-MI mortality and heart failure across one internal and two external cohorts, alongside the same ResNet architecture trained from scratch \cite{ref13}. Across all tasks, we use a common frozen-encoder evaluation protocol and exceed state-of-the-art performance with a 41-times smaller model.

\end{enumerate}

\section{Methods}

\subsection{Study design and patient data}

We investigated whether pretrained ECG representations retain clinically meaningful prognostic information across healthcare systems and whether incorporating patient-temporal structure during pretraining influences external transfer. Patient data for pretraining were obtained from the MIMIC-IV database, which contains electronic health records and physiological waveform data acquired at Beth Israel Deaconess Medical Center between 2008 and 2019 \cite{ref14}. The pretraining dataset comprised 800,035 10-second, 12-lead ECGs from 161,352 patients, each represented by 5,000 samples per lead at 500 Hz. For downstream evaluation, post-MI cohorts were constructed from MIMIC-IV and two independent HEEDB datasets \cite{ref15}, hereafter referred to as HEEDB-Boston and HEEDB-Atlanta. MIMIC-IV was used to train and validate all downstream components, whereas both HEEDB cohorts were held out for external evaluation. Downstream MIMIC-IV evaluation used training, validation and test sets (approximately 64\%, 16\% and 20\%), shared across encoders. All downstream MIMIC-IV patients were present in the pretraining archive; HEEDB therefore provides evaluation outside the PACE-ECG pretraining population. Cohort construction, partitioning, outcome definitions and preprocessing are described in Appendices A--B.

\subsection{PACE-ECG}

\begin{figure}[h]
    \centering
    \includegraphics[width=0.82\linewidth]{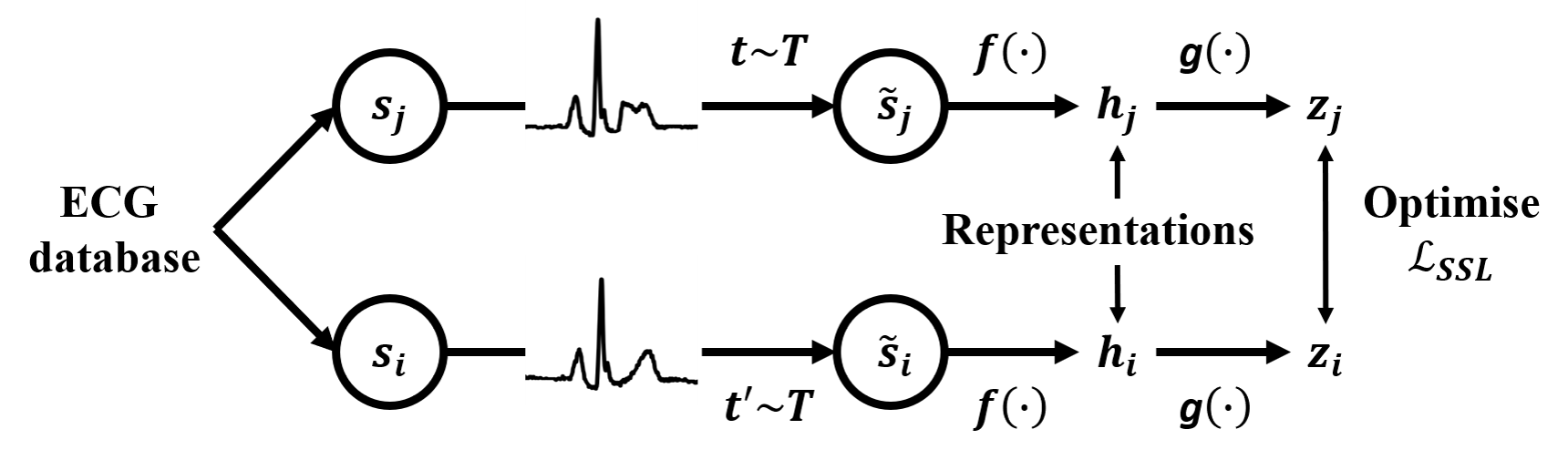}
    \caption{Proposed workflow. Two distinct ECGs are drawn from the database and augmented using stochastic transforms from the same family ($t\sim T$ and $t'\sim T$). A base encoder $f(\bullet)$ and a projection head $g(\bullet)$ are trained to maximize agreement between same-patient ECGs collected within 60 days. After training, the projection head is discarded, and the encoder is preserved for downstream tasks.}
    \label{fig:workflow}
\end{figure}

The proposed representation-learning framework (Figure~\ref{fig:workflow}) combines patient-temporal contrastive learning with clinically guided multitask supervision. The central objective is to exploit longitudinal structure in large ECG archives by encouraging representations of temporally neighbouring recordings from the same patient to remain close in embedding space.

\subsubsection{Patient-temporal contrastive learning}

Let ECG $i$ have patient identifier $s_i$ and acquisition time $t_i$. Two distinct recordings were eligible as a positive pair when they originated from the same patient and were acquired within a 60-day temporal window (Eq.~\ref{eq:pair}).

\begin{equation}
\mathcal{P} = \{(i,j): s_i=s_j,\ i\neq j,\ |t_i-t_j|\leq 60\}
\label{eq:pair}
\end{equation}

The 60-day window defines a local patient-specific neighbourhood for contrastive alignment. It is a fixed design choice, and future work should focus on establishing an empirical  optimum or a guarantee of physiological stability. Eligible pairs were enumerated independently for each patient and capped at 200 pairs per patient to limit over-representation of frequent attenders; patients with more eligible combinations were subsampled without replacement using a fixed seed, producing approximately 1.18 million pair rows. Each ECG in a pair then received an independent stochastic transformation. Unlike standard instance-level contrastive learning, positive pairs consist of two distinct longitudinal recordings, each independently augmented.

\subsubsection{Encoder and contrastive objective}

Each augmented ECG was processed by a ResNet1D encoder that maps a $12 \times 5{,}000$ input to a 128-dimensional representation $h$. This representation is then mapped to a contrastive space by a projection head $z=g(h)$ which is used only during pretraining and discarded for downstream evaluation. Minibatches contain 32 temporal pairs (64 ECGs), with the adjacent recording designated as the positive for each anchor. After L2 normalisation, cosine similarities are optimised using the normalised temperature-scaled cross-entropy loss (NT-Xent; Eq.~\ref{eq:ntxent}) where $N=64$. $p(k)$ denotes the patient-temporal partner of sample $k$ and temperature $\tau=0.3$ \cite{ref8}.

\begin{equation}
\mathcal{L}_{ssl} = -\frac{1}{N}\sum_{k=1}^{N}\log
\frac{\exp\!\left(\widetilde{z}_k^{\top}\widetilde{z}_{p(k)}/\tau\right)}
{\sum_{l\neq k}\exp\!\left(\widetilde{z}_k^{\top}\widetilde{z}_l/\tau\right)},
\qquad \widetilde{z}=\frac{z}{\lVert z\rVert_2}
\label{eq:ntxent}
\end{equation}

All non-designated recordings in the minibatch are treated as negatives. The objective encourages local, patient-conditioned similarity across repeated recordings while remaining robust to artificial signal perturbations. Accordingly, we refer to this method as patient-temporally structured contrastive pretraining.

\subsubsection{Clinically guided multitask supervision}

In addition to the patient-temporal contrastive objective, auxiliary physiological supervision is applied directly to the encoder representation. A seven-output regression head predicts ECG waveform measurements, and a two-output classification head predicts report-derived rhythm abnormalities. Applying the auxiliary tasks to the pre-projection representation encourages the transferable encoder itself to preserve morphology and rhythm information, while the projection head remains free to specialise for contrastive alignment. The supervised tasks are combined using learned homoscedastic uncertainty weights and jointly optimised with the contrastive objective. Exact target definitions, loss functions, augmentation parameters, architecture and optimisation hyperparameters are provided in Appendix C-D.

\subsection{Evaluation}

We compared PACE-ECG with three complementary benchmarks:
\begin{enumerate}
    \setlength{\itemsep}{0pt}
    \setlength{\parskip}{0pt}
    \setlength{\parsep}{0pt}
    \item DeepECG-SSL, a contemporary ECG foundation model. DeepECG-SSL combines Wav2Vec-based local representation learning, Contrastive Multi-Segment Coding and random lead masking. Its objectives learn local and global ECG structure. Because its pretraining data, architecture and supervision differ from PACE-ECG, this is a comparison of pretrained representations rather than a controlled test of an individual objective. We used the published W2V+CMSC+RLM checkpoint and its prescribed preprocessing, described in \cite{ref7}.
    \item Matched generic multitask SSL, trained with the same dataset and architecture as PACE-ECG and retaining the auxiliary supervision. The intended difference is the positive-pair definition: the generic control uses two independently augmented views of the same ECG rather than two distinct ECGs from the same patient within 60 days. This assesses the bounded same-patient pairing scheme relative to same-ECG augmentation within the combined objective; it does not isolate temporal restriction from patient identity.
    \item Supervised scratch, using the same encoder architecture but initialised randomly and trained jointly with the downstream prediction head using MIMIC-IV labels. This comparator tests whether pretraining improves the complete downstream solution and is not a pure frozen-representation comparison.
\end{enumerate}

The same ECGs were supplied to each pretrained encoder. For the primary foundation-model comparison, encoders were frozen and a nonlinear prediction head with a common hidden-layer design was trained using MIMIC-IV only, then applied unchanged to both HEEDB cohorts. All checkpoints were selected using MIMIC-IV validation performance only. Patient-level AUROC was the primary metric. Five downstream probe seeds were used to assess sensitivity to probe initialisation, and we report mean AUROC with the range across seeds in parentheses.

\section{Results}

Full results are reported in Tables 1 and 2. For mortality, the proposed model achieved the highest in-domain MIMIC-IV AUROC (0.788) compared with 0.773 for the DeepECG-SSL representation. The difference between the two models was sharper in HEEDB-Boston (0.769 versus 0.728), and even larger in HEEDB-Atlanta (0.732 versus 0.685). For heart failure, PACE-ECG slightly outperformed DeepECG-SSL in-domain (0.731 versus 0.703) but their performance difference was greater under external transfer: 0.779 versus 0.706 in Boston and 0.787 versus 0.709 in Atlanta. 

The matched ablation showed a consistent advantage for patient-temporal pairing over generic same-ECG SSL. For heart failure, patient-temporal pretraining increased AUROC by +0.0487 on MIMIC-IV (95\% paired CI 0.0258 to 0.0617), +0.0676 in HEEDB-Boston (0.0539 to 0.0715) and +0.0680 in HEEDB-Atlanta (0.0470 to 0.0791). Mortality gains were smaller: +0.0391 on MIMIC-IV (-0.0290 to 0.0505), +0.0494 in Boston (0.0368 to 0.0533) and +0.0591 in Atlanta (0.0385 to 0.0702). 

Thus, the clearest benefit of the temporal objective was observed under external transfer. All pretrained representations also produced higher AUROC than the supervised scratch model across both outcomes and all three cohorts, with the largest differences observed for external heart failure prediction. Because the scratch comparator updates the encoder using endpoint labels while the representation benchmark freezes pretrained encoders, we interpret this result as further evidence for the utility of pretraining.

\begin{table}[h]
\caption{All-cause mortality prediction post-MI. Values are mean AUROC across five downstream probe seeds, with the worst-to-best range in parentheses.}
\label{tab:mortality}
\centering
\normalsize
\setlength{\tabcolsep}{4.0pt}
\begin{tabular}{lccc}
\toprule
Model & MIMIC-IV & HEEDB Boston & HEEDB Atlanta \\
\midrule
PACE-ECG& \textbf{0.788 (0.786--0.789)}& \textbf{0.769 (0.767--0.770)}& \textbf{0.732 (0.728--0.735)}\\
SSL model & 0.749 (0.747--0.752) & 0.719 (0.715--0.722) & 0.672 (0.668--0.676) \\
DeepECG-SSL& 0.773 (0.772--0.774) & 0.728 (0.724--0.732) & 0.685 (0.684--0.686) \\
Trained from scratch & 0.747 (0.730--0.759) & 0.650 (0.628--0.690) & 0.659 (0.629--0.685) \\
\bottomrule
\end{tabular}
\end{table}

\begin{table}[h]
\caption{Incident heart failure prediction post-MI. Values are mean AUROC across five downstream probe seeds, with the worst-to-best range in parentheses.}
\label{tab:hf}
\centering
\normalsize
\setlength{\tabcolsep}{4.0pt}
\begin{tabular}{lccc}
\toprule
Model & MIMIC-IV & HEEDB Boston & HEEDB Atlanta \\
\midrule
PACE-ECG& \textbf{0.731 (0.725--0.734)}& \textbf{0.779 (0.777--0.781)}& \textbf{0.787 (0.784--0.790)}\\
SSL model & 0.683 (0.679--0.687) & 0.712 (0.708--0.715) & 0.720 (0.716--0.723) \\
DeepECG-SSL& 0.703 (0.702--0.705) & 0.706 (0.703--0.708) & 0.709 (0.702--0.715) \\
Trained from scratch & 0.651 (0.642--0.661) & 0.619 (0.606--0.631) & 0.668 (0.660--0.675) \\
\bottomrule
\end{tabular}
\end{table}

\section{Discussion}

This study shows that ECG foundation-model transfer cannot be evaluated on in-domain performance alone. In incident heart failure prediction PACE-ECG modestly outperformed DeepECG-SSL on MIMIC-IV, yet, its performance was considerably higher in both HEEDB-Boston and HEEDB-Atlanta. Mortality produced a slightly smaller improvement but a similar trend, with the proposed model outperforming DeepECG-SSL in-domain and externally. Taken together, these results suggest that representation quality depends on task and patient distribution. The generic multitask-SSL comparison supports the relevance of positive-pair construction within the evaluated combined objective. Replacing same-ECG views with temporally neighbouring same-patient recordings improved the reported performance across task--cohort comparisons. This supports the utility of the pairing scheme, but does not identify whether the gain arises from temporal proximity, stable patient characteristics or their interaction with auxiliary supervision. The comparison with DeepECG-SSL also places the results in the context of encoder size. DeepECG-SSL is a strong recent benchmark trained on more than 1.4 million ECGs and uses a 90.4-million-parameter encoder, whereas the proposed encoder contains 2.21 million parameters and was pretrained on 800,035 ECGs. Despite having approximately 41 times fewer encoder parameters, PACE-ECG outperformed DeepECG-SSL in all evaluated settings. This suggests that clinically aligned pretraining structure can be as important as model and dataset scale for a specific transfer setting. Our evaluation complements ECG interpretation benchmarks and prior prognostic transfer studies \cite{khattak2026} by focusing on mortality and incident heart failure after MI across the specified cohorts. The consistent advantage of pretrained models over supervised training from scratch, particularly on the external cohorts, further supports the value of learning reusable ECG representations before exposure to endpoint-specific labels.

This study has several limitations. Evaluation is restricted to two post-MI outcomes, three cohorts and one external foundation-model family, limiting generalisation to other tasks and architectures. The 60-day window was not compared with other intervals or unrestricted same-patient pairing; Future work should investigate the optimal duration of this time window. Same-patient alignment may retain stable identity, demographic or acquisition-related characteristics as well as clinically relevant physiology, and the present evaluation does not separate these mechanisms. Both pairing schemes retain auxiliary supervision, so their comparison does not quantify the independent contribution of those tasks or their interaction with pairing. Future work should test patient-temporal pretraining across additional architectures, including Transformer and joint-embedding predictive approaches, and across outcomes that require different temporal and physiological information.

\section{Conclusion}

We developed PACE-ECG, a patient-aware contrastive pretraining framework for ECG representation learning and evaluated its transfer to post-MI mortality and heart failure prediction across internal and external clinical cohorts. Within the evaluated multitask framework, using temporally neighbouring ECGs from the same patient as positives improved transfer relative to generic same-ECG SSL. PACE-ECG also outperformed a state-of-the-art ECG foundation model, 41-times the size, in all internal and external
 settings. These findings support two conclusions: the evaluated pretraining scheme influences downstream transfer, and in-domain benchmark performance alone is insufficient to establish the clinical generalisability of an ECG foundation representation. Broader multi-cohort evaluation on prognostic tasks should therefore form a central part of future biosignal foundation model assessment.

\clearpage
\renewcommand{\refname}{References}

\clearpage
\appendix
\setcounter{table}{0}
\renewcommand{\thetable}{\Alph{section}\arabic{table}}
\renewcommand{\theHtable}{\Alph{section}\arabic{table}}

\section{Baseline characteristics}

Table~\ref{tab:baseline} shows the baseline characteristics of the MI cohorts used across this study. Among the subsets, only MIMIC-IV MI was used to train the prediction heads for downstream binary classification. HEEDB Boston and HEEDB Atlanta were not used for PACE-ECG pretraining or for fitting or selecting the downstream prediction heads.

\paragraph{Patient partitioning and pretraining overlap.}
For each endpoint, MIMIC-IV patients were assigned to approximately 64\% training, 16\% validation and 20\% test partitions, stratified by the outcome label. Patient assignments were shared across encoders, with all recordings from a patient assigned to the same downstream partition. Validation data were used for downstream model selection, while the test partition was reserved for internal evaluation. These downstream partitions were patient-disjoint, but were not disjoint from representation pretraining: PACE-ECG was pretrained using the full MIMIC-IV ECG archive, including all downstream validation and test patients. The resulting 100\% patient overlap makes MIMIC-IV an in-domain evaluation in patients represented in the pretraining archive. The two HEEDB cohorts provide external evaluation outside that population; no external adaptation or model selection was performed.

\begin{table}[h]
\caption{Baseline characteristics of the post-MI cohorts used in this study.}
\label{tab:baseline}
\centering
\small
\begin{tabular}{lccc}
\toprule
Characteristic & MIMIC-IV MI cohort & HEEDB Boston & HEEDB Atlanta \\
\midrule
Patients & 8,494 & 10,222 & 1,855 \\
Mortality events & 3,074 & 1,393 & 329 \\
HF events & 2,407 & 1,644 & 664 \\
Age at MI, years, median (IQR) & 71.0 (61.0--81.0), & 65.8 (57.0--73.1) & 65.1 (54.9--73.0) \\
Female, n (\%) & 3,289 (38.8\%) & 3,659 (35.8\%) & 730 (39.4\%) \\
\bottomrule
\end{tabular}
\end{table}

\section{Disease definitions}

For downstream evaluation, MI was identified using standardised diagnosis codes from the international classification of diseases (ICD-10 I21; ICD-9 410). Separate endpoint-specific cohorts were constructed for all-cause mortality and incident heart failure. Patients with documented heart failure before the MI index were excluded from the incident-HF cohort. Mortality incorporated available in-hospital and post-discharge death information; heart failure was identified from subsequent diagnostic coding (ICD-9: 4280, 42820, 42821, 42822, 42823). Where multiple post-MI ECGs were available for the same patient, only one was retained. This was the first available ECG collected after MI. MIMIC-IV ECGs were 10 seconds long and sampled at 500 Hz. HEEDB ECGs acquired at 250 Hz were resampled to 500 Hz before evaluation.

\paragraph{Prediction time and observation period.}
The selected first post-MI ECG served as the patient-level predictor. Binary endpoints reflected deaths or HF diagnoses recorded during available follow-up, whose duration varied between patients and cohorts. The primary AUROC comparisons therefore describe discrimination of recorded outcomes over available observation, rather than performance at a common fixed prediction horizon. This discrimination-based evaluation does not establish calibration or clinical utility.

\section{Pretraining procedure and hyperparameters}

Pretraining ECGs were transposed to $12 \times 5{,}000$, and recordings with a different geometry were excluded. Non-finite samples were replaced with zero. Signals were standardised using fixed MIMIC-IV mean and standard deviation statistics applied identically to every lead and recording. Each patient contributed at most 200 temporal pairs; 1,178,560 pairs were processed per epoch in complete minibatches. All models and training were built on PyTorch 2.5.1 with CUDA 12.1 and Python 3.10.11.

\begin{figure}[t]
    \centering
    \includegraphics[width=0.54\linewidth]{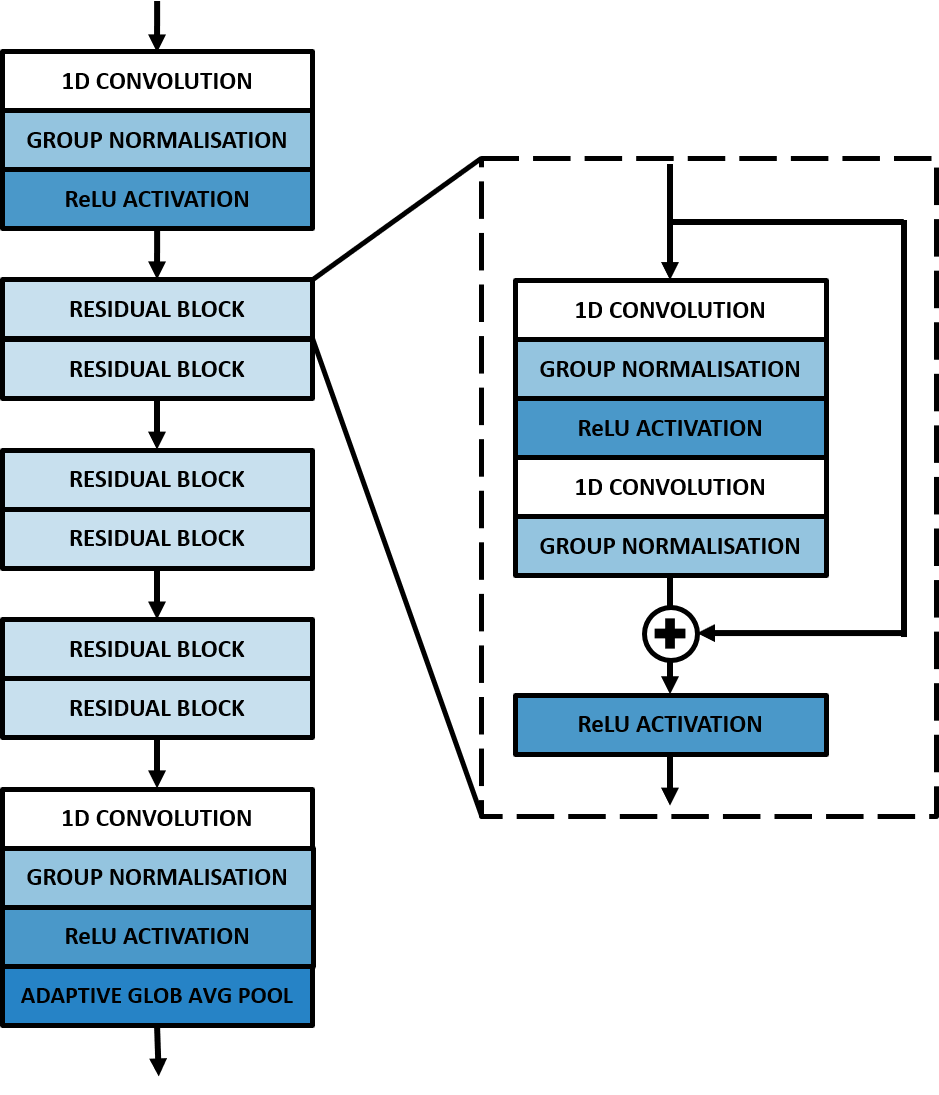}
    \caption{Architecture of the ResNet1D encoder used in the proposed model. Augmented 12-lead ECGs are processed through an initial convolutional stem followed by three stages of residual blocks, with the first block of Stages 2 and 3 using stride-2 downsampling and a projection shortcut. A final $1\times1$ convolution and adaptive global average pooling compress the signal into a 128-dimensional ECG representation.}
    \label{fig:resnet}
\end{figure}

The proposed ResNet1D (Figure~\ref{fig:resnet}) encoder uses base width 64, kernel size 7. It comprises an initial strided convolution, three residual stages, adaptive global average pooling, eight GroupNorm groups to avoid dependence on source-domain running activation statistics during transfer, and a 128-dimensional output (2,205,568 parameters). The contrastive projector is 128$\rightarrow$256$\rightarrow$128 neurons with LayerNorm and was discarded after pretraining. Augmentations used during contrastive pretraining were Gaussian noise ($\sigma$=0.01, always), global amplitude scaling (uniformly sampled between 0.9 and 1.1), circular temporal shift ($p$=0.8; up to $\pm$0.5 s), two 0.8-s time masks ($p$=0.5), dropout of 1-3 leads ($p$=0.3), and baseline wander ($p$=0.5; 5\% of per-ECG RMS). Training used AdamW (learning rate $1 \times 10^{-3}$; PyTorch default weight decay 0.01), batches of 32 pairs/64 ECGs, no learning-rate scheduler or warm-up, and CUDA automatic mixed precision on an NVIDIA RTX 4090. Losses were evaluated in float32 for numerical stability. The run was configured for 80 epochs and produced complete checkpoints through epoch 50; no implemented early-stopping criterion was used. For both mortality and heart failure prediction, we used the same two-layer multilayer perceptron probe. This included $d$$\rightarrow$128$\rightarrow$128$\rightarrow$1 where $d$ is the flattened shape of the ECG representation. $d=128$ for the proposed, generic-SSL, and randomly initialized ResNet encoders, and $d=768$ for DeepECG-SSL. ReLU activation and dropout ($p$=0.2) were included between the dense layers. The final layer produced a single logit, optimized using binary cross-entropy with logits; with sigmoid applied to obtain predicted probabilities. Separate probes were trained for mortality and heart failure, while the pretrained encoders remained frozen in the primary experiments.

\section{Supervised auxiliary tasks}

The seven continuous auxiliary targets used to regularise the contrastive task were P-wave duration, QRS duration, ST-T duration, P-wave axis, QRS axis, T-wave axis and RR interval. The two binary targets were atrial fibrillation and an MI-related report abnormality indicator derived from ECG-report text; the latter is broader than the downstream MI phenotype and includes ischaemic and ST-T terminology. Regression used per-target mean squared error (Eq.~\ref{eq:reg}), while binary classification used binary cross-entropy with logits (Eq.~\ref{eq:cls}). Per-target homoscedastic log-variance parameters were learned jointly following Kendall et al. \cite{ref16}, producing the supervised objective in Eq.~\ref{eq:sup}.

\begin{equation}
\mathcal{L}_{r}^{reg}=\frac{1}{B}\sum_{b=1}^{B}\left(\widehat{y}_{br}-y_{br}\right)^2
\label{eq:reg}
\end{equation}

\begin{equation}
\mathcal{L}_{m}^{cls}=-\frac{1}{B}\sum_{b=1}^{B}\left[y_{bm}\log\sigma(a_{bm})+(1-y_{bm})\log\left(1-\sigma(a_{bm})\right)\right]
\label{eq:cls}
\end{equation}

\begin{equation}
\mathcal{L}_{\sup}=\sum_r\left(\exp(-s_r)\,\mathcal{L}_{r}^{reg}+s_r\right)+\sum_m\left(\exp(-u_m)\,\mathcal{L}_{m}^{cls}+u_m\right)
\label{eq:sup}
\end{equation}

The learned uncertainty terms adapt the contribution of targets with different numerical scales and noise levels. The contrastive and supervised branches were trained jointly using the objective in Eq.~\ref{eq:joint}. The encoder receives gradients from both branches; the projection head receives only contrastive gradients, while the auxiliary heads and learned uncertainty parameters receive only supervised gradients.

\begin{equation}
\mathcal{L}=2.0\bullet\mathcal{L}_{SSL}+0.2\bullet\mathcal{L}_{\sup}
\label{eq:joint}
\end{equation}

\end{document}